\documentclass{article} 
\usepackage{iclr2025_conference,times}

\usepackage{amsmath,amsfonts,bm}

\def\eqref#1{equation~\ref{#1}}

\def\1{\bm{1}}

\DeclareMathAlphabet{\mathsfit}{\encodingdefault}{\sfdefault}{m}{sl}
\SetMathAlphabet{\mathsfit}{bold}{\encodingdefault}{\sfdefault}{bx}{n}

\usepackage{hyperref}
\usepackage{url}
\usepackage{graphicx}
\usepackage{gensymb}
\usepackage{textcomp}
\usepackage{subcaption}
\usepackage{makecell}
\usepackage{booktabs}
\usepackage{algorithm2e, algpseudocode}
\usepackage{wrapfig}
\RestyleAlgo{ruled}
\usepackage{amsmath}
\usepackage{titlesec}

\titlespacing*{\section}{0pt}{1em}{0ex}

\title{HoloGarment: 360\degree\ Novel View Synthesis of In-the-Wild Garments}

\iclrfinalcopy

\author{Johanna Karras$^{1,2}$, Yingwei Li$^{2}$, Yasamin Jafarian$^{2}$, Ira Kemelmacher-Shlizerman$^{1,2}$  \\
$^{1}$University of Washington, $^{2}$Google \\
}

\begin{document}

\maketitle

\begin{figure}[h]
\begin{center}
   \includegraphics[width=\textwidth]{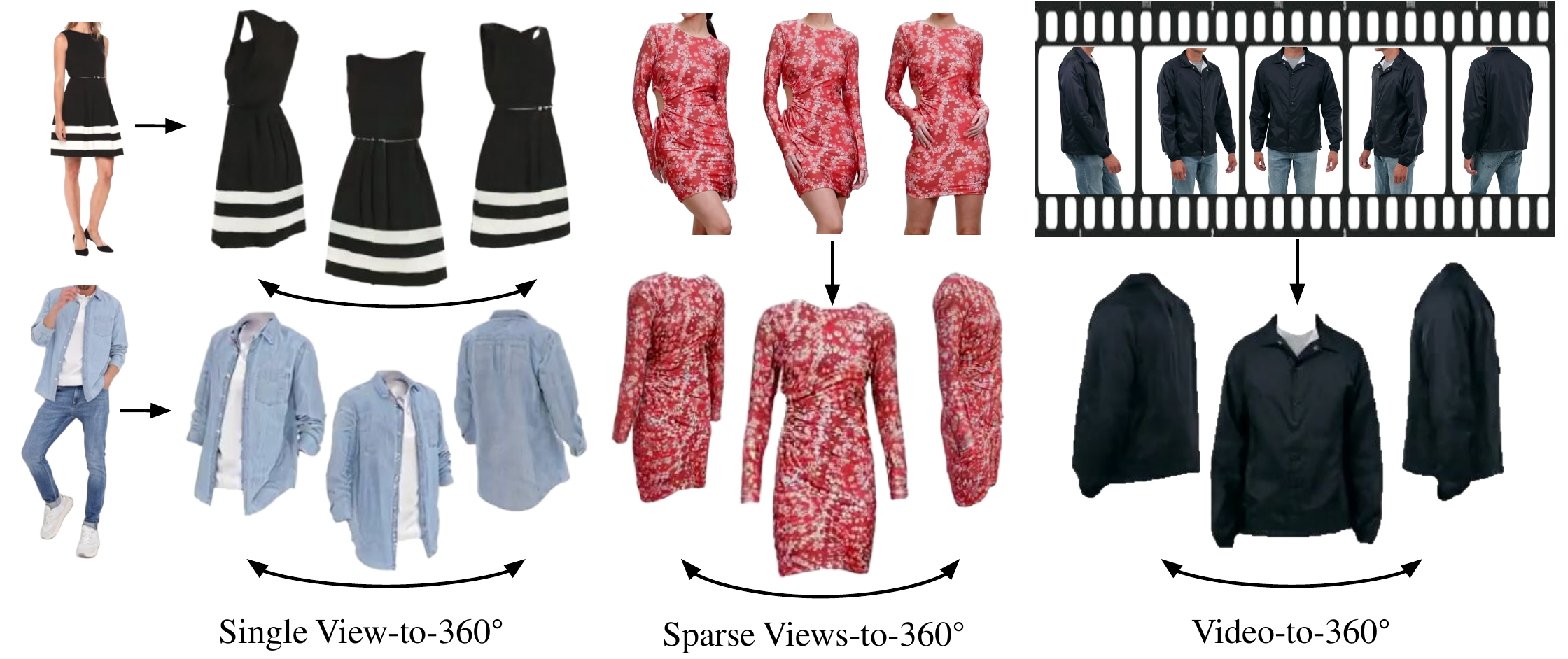}
   \caption{HoloGarment enables 360\degree\ novel view synthesis of real-world garments in images and videos.}
   \label{fig:teaser}
\end{center}
\end{figure}

\begin{abstract}

Novel view synthesis (NVS) of in-the-wild garments is a challenging task due significant occlusions, complex human poses, and cloth deformations. Prior methods rely on synthetic 3D training data consisting of mostly unoccluded and static objects, leading to poor generalization on real-world clothing. In this paper, we propose HoloGarment (\textbf{Holo}gram-\textbf{Garment}), a method that takes 1-3 images or a continuous video of a person wearing a garment and generates 360\degree\ novel views of the garment in a canonical pose. Our key insight is to bridge the domain gap between real and synthetic data with a novel implicit training paradigm leveraging a combination of large-scale real video data and small-scale synthetic 3D data to optimize a shared garment embedding space. During inference, the shared embedding space further enables dynamic video-to-360\degree\ NVS through the construction of a garment “atlas” representation by finetuning a garment embedding on a specific real-world video. The atlas captures garment-specific geometry and texture across all viewpoints, independent of body pose or motion. Extensive experiments show that HoloGarment achieves state-of-the-art performance on NVS of in-the-wild garments from images and videos. Notably, our method robustly handles challenging real-world artifacts -- such as wrinkling, pose variation, and occlusion -- while maintaining photorealism, view consistency, fine texture details, and accurate geometry. Visit our \href{https://johannakarras.github.io/HoloGarment/}{project page} for additional results: https://johannakarras.github.io/HoloGarment/

\end{abstract}

\section{Introduction}
\label{sec:intro}

The rise in online retail, virtual try-on, and digital fashion design is driving the demand for high-quality digital garment visualizations. While images and videos offer glimpses of a garment from various angles, they fall short of providing a \textbf{full 360\degree\ representation -- one that is independent of the wearer and free from occlusions or wrinkles}. Manually acquiring such 360\degree\ views of real garments is highly impractical, since capturing dense multi-view data is costly and time-consuming. 
Therefore, there is strong interest in discovering an automatic method to generate high-quality, high-fidelity novel views of garments from 2D images and videos.

This is a challenging task, as real-world garments are inherently complex, containing \textbf{deformations, occlusions, and pose variations} when worn. Existing methods for novel view synthesis (NVS) of general objects~\citep{cat3d, trellis} only handle a fixed number of input views and are constrained to \textbf{static and unoccluded objects} in fixed poses. As a result, these approaches perform poorly on real-world garments and do not handle an arbitrary number of input views, such as from a video. While some methods handle finetuning on video~\citep{vid2vid,neural_talking_heads}, they are prone to overfitting to the shape and appearance of the subject in the input frames. As a result, finetuning with existing methods on a dynamic garment video containing self-occlusions and deformations fails to generate unoccluded, static novel views of the standalone garment.

Another challenge is that currently available 3D garment datasets~\citep{dresscode, diffavatar, deepfashion3d} are synthetic, which limits their size, diversity, and realism. On the other hand, real 2D garment data, images and videos, are abundantly available online, but are missing ground-truth 3D representations. As such, past works~\citep{deepcloth, cloth2tex, garment3dgen, bang_patterns, spnet, neuraltailor, dresscode, TEXTure} often leverage purely synthetic garment data, tending to overfit to simplistic shapes and patterns and generalizing poorly to diverse garment shapes and textures in the wild. 

To overcome these limitations, we seek to answer the questions: (1) Is it possible to train a real-world garment NVS model by leveraging abundant real-world 2D data, even in the absence of paired ground-truth 3D assets? (2) In contrast to sparse-frame conditioning, can dense frames from a dynamic video sequences be leveraged to learn a robust and geometrically accurate garment representation for NVS?


In this paper, we propose \textbf{HoloGarment, a video diffusion model for garment NVS from images and videos of in-the-wild dressed humans}. Our key insight is \textbf{a novel implicit training paradigm}, where two or more distinct training tasks indirectly train a model to perform the target task for which ground truth data is not available. Using a combination of real 2D data and synthetic 3D assets, our method learns a shared garment embedding space between both domains that enables real-world garment novel view synthesis. In doing so, we bypass the limitations of synthetic-only 3D datasets to handle challenging real-world garment images and videos. Furthermore, \textbf{we introduce the notion of a garment ``atlas"}, a finetuned garment embedding optimized on a specific dynamic video featuring a person wearing the garment. The ``atlas" bridges the gap between finetuning (2D) and inference (3D) modalities, \textbf{enabling the novel task of \textit{video}-to-NVS generation}, as well as eliminates the need for arbitrary input view selection.

Our experiments showcase HoloGarment's capability to generate high-quality, high-fidelity 360\degree\ novel views across a variety of garment types, including tops, dresses, jackets, rompers, and pants, even those containing occlusions, pose variations, and deformations. We further quantitatively and quantitatively demonstrate that our method \textbf{achieves state-of-the-art results} compared to related methods.

\section{Related Work}

\textbf{Novel View Synthesis with Diffusion Models} Novel view synthesis refers to generating novel object views from limited observations, such as images. A common framework involves training a diffusion model with 3D datasets. Several methods adopt this paradigm by finetuning pretrained text-to-image diffusion models~\citep{cat3d, mvdream, imagedream, zero123, zero123++, syncdreamer} or video diffusion models~\citep{stable_virtual_camera, motionctrl, vivid-1-to-3}. However, these approaches are constrained by their reliance on 3D data, which limits their ability to handle real-world images effectively. Several works have explored using 2D diffusion priors to enhance 3D consistency~\citep{ dreamfusion, magic3d, mvdream}, but do not tackle cases where input views contain incomplete information (i.e. occlusions) or inconsistencies (i.e. deformations, pose changes). Therefore, existing NVS approaches cannot handle diverse garments in complex and dynamic real-world scenarios. In this work, we address this challenging case directly by training a video diffusion model implicitly on real 2D videos and synthetic 3D assets. This training strategy enables our method to robustly generate consistent multi-view images of real-world garments.

    
\textbf{3D Garment Reconstruction} Related to the task of garment novel view synthesis is 3D garment reconstruction, which aims to recover the 3D geometry of a garment in an image. One avenue of garment 3D reconstruction methods explores the estimation of 2D sewing patterns~\citep{bang_patterns, spnet, neuraltailor, dresscode, sewformer}, which provide a foundation for realistic garment modeling by leveraging flat patterns that can be draped into the person's 3D structures. However, these methods often focus solely on representing geometry, neglecting to preserve texture details. Other recent methods focus on texture estimation by utilizing template garment meshes to achieve better realism in garment representation~\citep{deepcloth, TEXTure, garment3dgen, cloth2tex}. 
A major limitation of these methods is their reliance on limited synthetic 3D garment datasets, including DressCode~\citep{dresscode} and GarverseLOD~\citep{garverselod}, which also do not include ground-truth textures. As a result, these methods do not generalize well to  real-world garment inputs. In contrast to these methods, our approach eliminates the reliance on purely synthetic data, input meshes, and complex templates. 

\textbf{Subject-Specific Finetuning} Subject-specific finetuning, or personalization, refers to finetuning a pre-trained generative model to produce outputs of a specific subject. Notably, DreamBooth~\citep{dreambooth} customizes text-to-image diffusion models to generate images of a specific subject using a specialized token. Other works~\citep{vid2vid,neural_talking_heads} customize generative video models to produce videos of a specific subject by training in a few-shot manner. Similarly, this paradigm has been extended for human identity preservation in various other diffusion model applications, such as try-on~\citep{mixmatch} and human animation~\citep{dreampose}. To the extent of our knowledge, subject-specific finetuning has not been applied to garment identity specifically, which comes with unique challenges. One specific limitation is that current subject-finetuning methods tend to overfit to the pose and shape of the target subject. As such, existing methods will replicate the motion, occlusions, deformations, and wrinkling of a dynamic garment video that is used for finetuning. This is undesirable for synthesizing a static video of a dynamic garment in an unoccluded, canonical A-pose. We address this by explicitly disentangling animation and spin motion via split temporal blocks in our network, while still sharing the same garment appearance encoder. As a result, we are able to finetune a garment-specific embedding on a dynamic garment video and still be able to generate 360\degree\ novel views of the garment in a static pose, without overfitting to the original dynamic motion.



    
    
    

\section{Preliminaries}

In this section, we provide background on diffusion models and transformer diffusion models, including the video transformer diffusion model~\citep{fashion-vdm}, which is the backbone of our method.

\textbf{Diffusion Models:} Diffusion models are a class of generative models capable of synthesizing high-fidelity data, particularly images and videos~\citep{sohl2015deep, songerman2019,ddpm,ddim, dhariwal2021diffusion}. In the forward process, the data (e.g. image or video) is transformed incrementally into pure Gaussian noise over a discrete number of steps. Then, a diffusion model (typically a UNet) is trained to predict the reverse process, which iteratively denoises the Gaussian noise back into a clean data sample. To be precise, at timestep $t$, diffusion model $\epsilon_{\theta}$ with parameters $\theta$ predicts noise $\hat{\epsilon}_t$ added to the noisy data sample $z_{t}$. With conditioning signals $c$, one diffusion timestep is defined as:

    \begin{equation}
        \hat{\epsilon}_t = \epsilon_{\theta}(z_t, t, c)
    \end{equation}
    \label{eq:dm}

From the predicted noise, the denoised data sample $\hat{z}_{t-1}$ can be estimated. The diffusion model is optimized by the following objective function:

    \begin{equation}
        \mathcal{L} = ||\epsilon_t - \epsilon_{\theta}(z_t, t, c)||_2^2
    \end{equation}
    \label{eq:loss}
    
\textbf{Video Diffusion Transformer Models:}  While conventional diffusion models often leverage a U-Net backbone, the diffusion transformer (DiT) \citep{DiT} model replaces this with a Transformer architecture, leading to superior scalability and performance. 
Fashion-VDM~\citep{fashion-vdm} extends DiT into a video model with temporal blocks (e.g. 3D convolutions and temporal attention layers) and progressive temporal training. Paired with parallel UNet encoders to disentangle person and garment conditioning signals~\citep{tryondiffusion}, Fashion-VDM achieves superior performance for video try-on.

\section{Method}


Given 1-3 images $I_g$ of a garment $g$ and a driving pose sequence $J=(J_{2D}, J_{3D})$ represented in 2D and 3D, HoloGarment generates novel garment views $\hat{V}_g$ following the driving poses. In this section, we introduce our model architecture (\ref{ssec:hologarment}) and implicit training strategy with real and synthetic garment data (\ref{ssec:method-implicit-training}). Then, we describe how this unlocks image(s)-to-360\degree\ novel view synthesis (\ref{ssec:method-image-to-360}), as well as video-to-360\degree\ novel view synthesis capabilities via finetuning a garment ``atlas" (\ref{ssec:atlas}). 

    




\begin{figure}[!tbp]
  \centering
  \includegraphics[width=\linewidth]{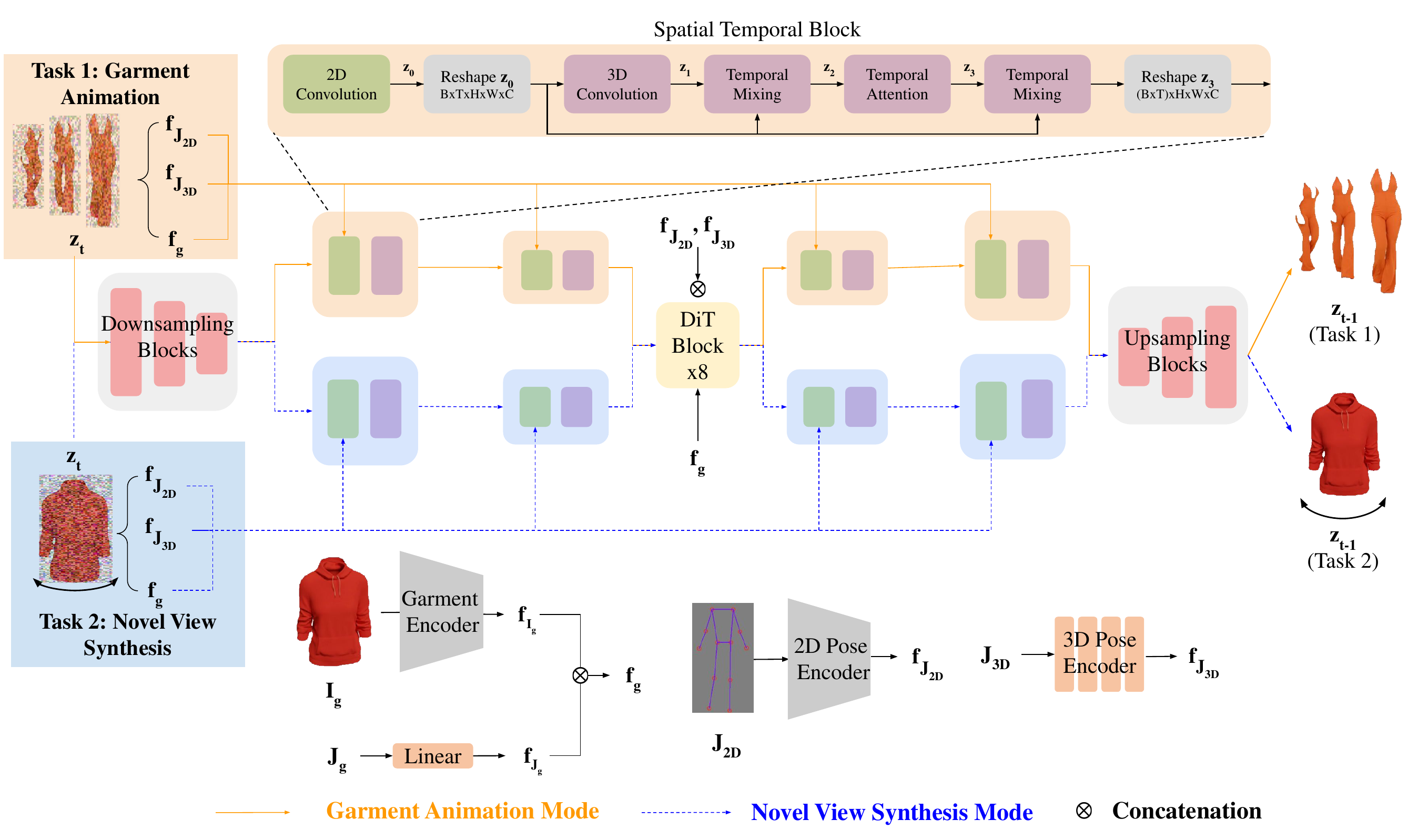}
  \caption{\textbf{Architecture.} Given an input garment image $I_g$ and a driving pose sequence $(J_{2D}, J_{3D})$, HoloGarment generates a video of the garment following the driving poses. The VDM operates in two distinct modes -- animation (solid orange line) and novel view synthesis (dotted blue line) -- depending on the input pose sequence. The UNet contains two separate sets of spatial temporal blocks~\citep{align-your-latents} -- one set for task 1 (orange) and one set for task 2 (blue). Only one branch is active at a time. All other spatial layers are shared between the two tasks. The noisy input frames $z_t$, input garment $I_g$, and driving 2D poses $J_{2D}$ are encoded by separate UNet encoders into $f_z, f_{I_g}, f_{J_{2D}}$, while the garment pose $J_g$ and 3D driving pose sequence $J_{3D}$ are encoded into $f_{J_g}, f_{J_{3D}}$ by a single linear and 4 dense layers, respectively. Garment features $f_{I_g}, f_{J_g}$ are concatenated into $f_g$. Inside of the DiT blocks, the garment features $f_g$ are cross-attended as keys and values with $f_z$, while $f_{J_{2D}}, f_{J_{3D}}$ are concatenated with $f_z$. The final noisy latent features $f_z$ are decoded by a UNet decoder to obtain $z_{t-1}$.}
  \label{fig:architecture}
\end{figure}

\subsection{HoloGarment}\label{ssec:hologarment}

At its core, our method consists of an image- and pose-conditioned video diffusion model (VDM) with trainable parameters $\theta$. Its architecture (Figure~\ref{fig:architecture}) builds upon the video transformer diffusion model proposed in Fashion-VDM~\citep{fashion-vdm}. However, it does not include any person image representation and the driving poses are encoded in both 2D and 3D. Our VDM additionally implements two identical sets of temporal blocks to separately handle video motion and 3D spin motion. We describe these adaptations in further detail below. Additional architecture and implementation details are provided in the supplementary material.

\textbf{Garment and Pose Conditioning:} Given a noisy video $z_t$ at diffusion timestep $t$, a UNet encoder $\mathcal{E}_z$ encodes $z_t$ into features $f_z$. Similarly, the conditional inputs -- garment image(s) $I_g$, garment pose(s) $J_g$, and driving poses $(J_{2D}, J_{3D})$ -- are encoded by their respective encoders $\mathcal{E}_{*}$ to compute features: $\mathcal{E}_{I_g}(I_g) = f_{I_g},\mathcal{E}_{J_g}(J_g) = f_{J_g}, \mathcal{E}_{J_{2D}}(J_{2D}) = f_{j_{2D}}, \mathcal{E}_{J_{3D}}(J_{3D}) = f_{J_{3D}}$. Mathematically, at timestep $t$, the VDM performs one denoising step to recover noise $\hat{\epsilon}_t$.

    \begin{equation}
        \hat{\epsilon}_t = \text{VDM}_{\theta}(z_t, t, f_g, f_j)
    \end{equation}
    \label{eq:vdm}

where $f_g = f_{I_g} \oplus f_{J_g}$ and $f_j = f_{J_{2D}} \oplus f_{J_{3D}}$. Here $\oplus$ indicates concatenation.

Inside the VDM, these conditional features are processed in the VDM via transformer blocks (DiT~\citep{DiT}). Pose features $f_j$ are spatially-aligned with $f_z$, so they are concatenated channel-wise before self-attention. Meanwhile, the non-spatially aligned garment features $f_g$ and noisy video features $f_z$ are cross-attended. In this manner, the garment features are implicitly warped to their target locations according to the driving poses~\citep{tryondiffusion}. 

\begin{figure}[!tbp]
  \centering
  \includegraphics[width=\textwidth]{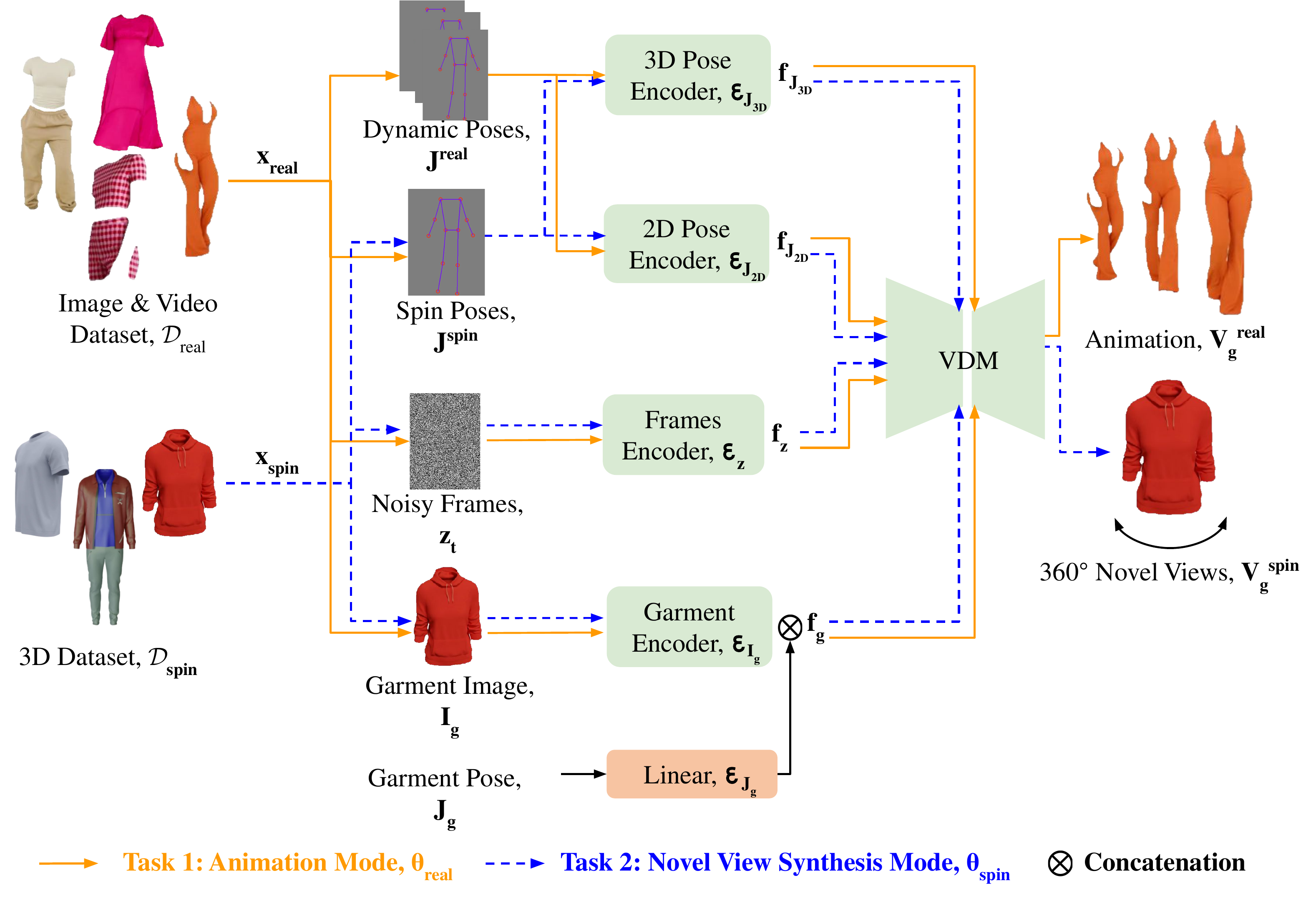}
  \caption{\textbf{Implicit Training Paradigm} A video diffusion model (VDM) is trained to generate either a garment animation given dynamic driving poses (solid orange path) or 360\degree\ novel views given static spin driving poses (blue dotted path). To better disentangle output motion styles, the VDM operates with parameters $\theta_{\text{real}}$ for animation and with parameters $\theta_{\text{spin}}$ for novel view synthesis. Parameters $\theta_{\text{real}}$ and $\theta_{\text{spin}}$ are shared except for their temporal blocks, which are distinct. By training on both real 2D data and synthetic 3D data, the VDM implicitly learns to generate canonical 360\degree\ novel views from 1-3 input images of real-world garments, without paired real-world 3D data.}
  \label{fig:implicit-training}
\end{figure}

\textbf{Disjoint Temporal Blocks:} To effectively disentangle dynamic and static spin motions, we implement our VDM with two identical, disjoint sets of temporal blocks. Each set consists of 3D-convolution, temporal attention, and temporal mixing blocks~\citep{align-your-latents}. One set is trained only on batches of \textit{real}-world images and videos $\mathcal{D}_{\text{real}}$ and the other is trained only on batches of static \textit{spin} renderings of 3D garment assets $\mathcal{D}_{\text{spin}}$. To specify which temporal blocks are activated, we refer the VDM parameters as $\theta_{\text{real}}$ when the $\mathcal{D}_{\text{real}}$-specific temporal blocks are activated and $\theta_{\text{spin}}$ when the $\mathcal{D}_{\text{spin}}$-specific temporal blocks are activated. Note that all non-temporal model parameters are shared between $\theta_{\text{real}}$ and $\theta_{\text{spin}}$. Disjoint temporal blocks allow the VDM to better synthesize motion in two modes -- dynamic or static spin -- depending on the which set of temporal blocks are activated.

\subsection{Implicit Training with Real and Synthetic Data}\label{ssec:method-implicit-training}
    
    To learn photorealistic garment novel view synthesis without real-world paired data, we formulate a novel paradigm of implicit training to learn a garment embedding space $F_G$ from both real 2D and synthetic 3D data. Implicit training leverages two or more related tasks for jointly training a model to perform a desired task. In contrast to $\textit{joint}$ training~\citep{VDM}, where ground-truth task-specific data is supplemented with similar, non-task-specific data, an $\textit{implicit}$ training strategy leverages solely non-task-specific data to learn the target task.  
    
    We postulate that if each task in implicit training offers part of the necessary learning for the desired task, then these tasks together can provide the full scope of necessary learning. In this case, the desired task is 3D-consistent novel view synthesis from real-world garment images, and the training tasks are (1) garment animation using real image and video data $\mathcal{D}_{\text{real}}$ and (2) novel view synthesis using synthetic 3D data $\mathcal{D}_{\text{spin}}$. Garment animation with real data trains the model to handle the desired \textit{input style} -- diverse, real-world garments, even under challenging conditions, like occlusions and wrinkling. Novel view synthesis with synthetic 3D data trains the model to generate the desired \textit{output style} -- unoccluded, static 360\degree\ views (spin videos) of garments.
    
    As shown in Figure~\ref{fig:implicit-training}, we train our VDM by alternating batches $x_{*}$ from both datasets:
    
    \begin{equation}
        x_{\text{real}}=(V_g^{\text{real}}, I_g^{\text{real}}, J_{g}, J^{\text{real}}) \sim \mathcal{D}_{\text{real}}
    \end{equation}
    \vspace{1em}
    \begin{equation}
         x_{\text{spin}}=(V_g^{\text{spin}}, I_g^{\text{spin}}, J_{g}, J^{\text{spin}}) \sim \mathcal{D}_{\text{spin}}
    \end{equation}
    \vspace{1em}
    
    During training, the VDM trains different temporal parameters for handling dynamic motion and static spin motion. For dynamic batches $x_{\text{real}}$, the VDM operates with $\theta_{\text{real}}$ and for spin batches $x_{\text{spin}}$, the VDM operates with $\theta_{\text{spin}}$ (Section~\ref{ssec:hologarment}). In this way, disjoint sets of temporal blocks are separately optimized for the different motion styles.
    
    After encoding the conditional inputs,
    
    \begin{equation}
     \label{eq:two-vdm-modes}
        \hat{\epsilon}_t = \begin{cases} 
              \text{VDM}_{\theta_{\text{real}}}(z_t, t, f_g^{\text{real}}, f_j^{\text{real}}) & x_{\text{real}} \sim \mathcal{D}_{\text{real}} \\
              \text{VDM}_{\theta_{\text{spin}}}(z_t, t, f_g^\text{spin}, f_j^{\text{spin}}) & x_{\text{spin}} \sim \mathcal{D}_{\text{spin}}
           \end{cases}
    \end{equation}
   \vspace{1em}

    Recall from Section~\ref{ssec:hologarment}, $f_g^{\text{real}} = \mathcal{E}_g(I_g^{\text{real}})$ and $f_g^{\text{spin}} = \mathcal{E}_g(I_g^{\text{spin}})$. Let $F_G$ be the garment encoder's ($\mathcal{E}_g$) embedding space for all garments $g \sim G$. Then,
    
    \begin{equation}
    \label{eq:shared-garment-embedding}
        f_g^{\text{real}}, f_g^{\text{spin}} \sim F_G
    \end{equation}
    \vspace{0.5em}
    
    Thus, real and synthetic garment embeddings share an embedding space $F_G$, which is optimized for both sets of model parameters ($\theta_{\text{real}}$ or $\theta_{\text{spin}}$). This implies that both input garment styles (real and synthetic) are compatible with both output motion styles (dynamic or static spin). Critically, this property enables HoloGarment to mix and match input garment styles with output motion styles. 
    
    
    

\subsection{Real-World Garment Image(s)-to-360\degree\ Garment}\label{ssec:method-image-to-360}

Our implicit training approach enables us to train a robust garment embedding space $F_G$ on diverse, large-scale garment video data that is also compatible with the novel view synthesis task. As a result, we can accomplish the desired implicit task of \textit{real} image-to-360\degree\ novel view synthesis. Given a real garment image $I_g^{\text{real}}$ (task 1) and static spin pose sequence in a canonical A-pose $J^{\text{spin}}$ (task 2), HoloGarment generates static 360\degree\ novel views of the input garment. During inference, the VDM operates with parameters $\theta_{\text{spin}}$ and denoises noisy frames $z_t$  via iterative noise prediction:

    \begin{equation}
        \hat{\epsilon}_t = \text{VDM}_{\theta_{\text{spin}}}(z_t, t, f_g^{\text{real}}, f_j^{\text{spin}})
    \end{equation}
    \vspace{1em}
    
    Recall that $f_g^{\text{real}} \sim F_G$ (Eq.~\ref{eq:shared-garment-embedding}) is compatible with both $\theta_{\text{real}}$ and $\theta_{\text{spin}}$. Therefore, our trained video diffusion model $\text{VDM}_{\theta_{\text{spin}}}$ generates consistent novel views of the real-world input garment style of task 1 in the output motion style of task 2: static, a-posed, and without occlusions, deformations, or wrinkling.

    
\subsection{Video-to-360\degree\ Garment via Atlas Finetuning}\label{ssec:atlas}

    \begin{wrapfigure}{r}{0.6\textwidth}
  \centering
  \vspace{-2em}
  \includegraphics[width=\linewidth]{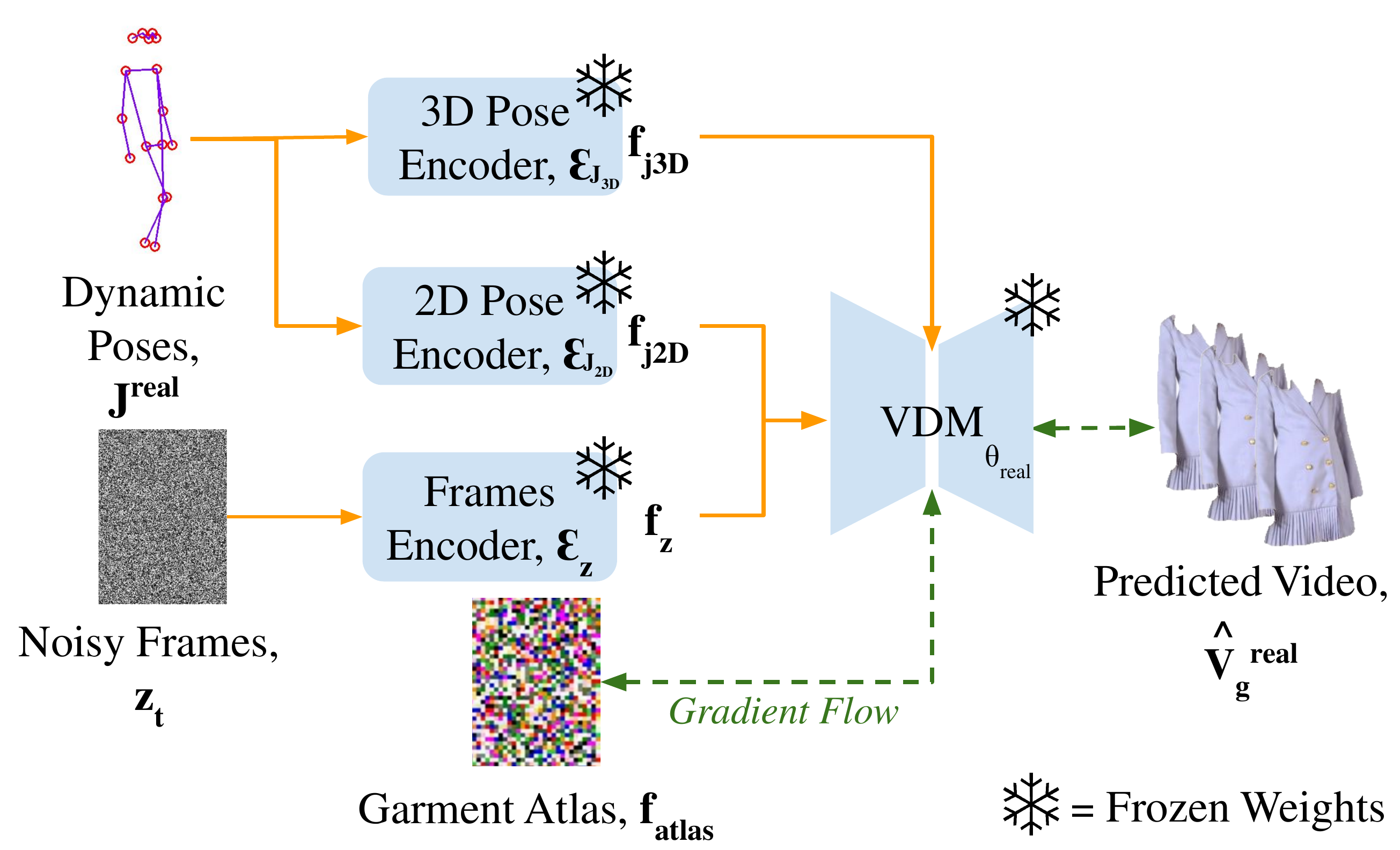}
  \vspace{-2.2em}
  \caption{\textbf{Garment Atlas Finetuning} HoloGarment enables video-to-NVS by finetuning a garment-specific embedding, or ``atlas", on a real-world video. By utilizing this atlas during inference, HoloGarment generates photorealistic 360\degree\ novel views of the garment.}
  \label{fig:atlas-finetuning}
  \vspace{-1em}
\end{wrapfigure}
    
    As a result of our implicit training approach (\ref{ssec:method-implicit-training}), the VDM's garment embedding space $F_G$ is shared between the 2D video animation and 3D novel view synthesis tasks. This enables another capability: finetuning a single garment embedding $f_g \sim F_G$ on a dynamic video (task 1) to run novel view synthesis (task 2). We call this finetuned garment-specific embedding the garment ``atlas", $f_{\text{atlas}}$, which can be leveraged for both tasks (Eq.~\ref{eq:shared-garment-embedding}). Note that this is in contrast with earlier subject finetuning methods~\citep{dreambooth}, which finetune a model on the same task (text-to-image) as during inference (text-to-image).
    
    The garment atlas finetuning strategy is shown in Figure~\ref{fig:atlas-finetuning}. Initially, $f_{\text{atlas}}$ is randomly initialized with same the shape as $f_g$, the non-optimized garment embeddings. Then, after freezing all other model parameters $\theta_{\text{real}}$, $f_{\text{atlas}}$ is finetuned on a specific garment video $V_g$ for $M$ iterations. Refer to Algorithm~\ref{alg:atlas-finetuning} for additional details.
    
    During inference, $f_{\text{atlas}}$ replaces the original garment embeddings $f_g$,
    
    \begin{equation}
        \hat{\epsilon}_t = \text{VDM}_{\theta_{\text{spin}}}(z_t, t, f_{\text{atlas}}, f_j^{\text{spin}})
    \end{equation}
    \vspace{1em}
    
    \SetKwInput{KwIn}{Input}
\SetKwInput{KwOut}{Output}
\SetKwInput{KwInit}{Initialize}

\begin{algorithm}[!h]
  \caption{Garment Atlas Finetuning on a Dynamic Garment Video}\label{alg:atlas-finetuning}
  \DontPrintSemicolon

  \KwIn{$V_g^\text{real}$: Input dynamic garment video}
  \KwOut{$f_{\text{atlas}}$: The finetuned garment embedding.}
  
  \KwInit{\\
    \Indp
    $\bullet$ Freeze all parameters $\theta$\\
    \Indp
    $\bullet$ $f_\text{atlas} \gets \text{random embedding of shape } f_g$
    \Indm
}

  \For{$i = 1$ to $M$}{
    Sample frames $v_g \sim V_g^\text{real}$\;
    Sample timestep $t \sim \texttt{Uniform}(1, T)$\;
    Sample noise $\epsilon_t \sim \texttt{Gaussian}(0, I)$\;
    Compute poses $(J_{2D}^{\text{real}}, J_{3D}^{\text{real}})$ from $v_g$\;
    Compute pose embeddings $f_j^{\text{real}}$\;
    $z_t \gets \texttt{AddNoise}(v_g, \epsilon_t, t)$ \Comment{Get noisy frames}\;
    $\hat{\epsilon_t} \gets \text{VDM}_{\theta_{\text{real}}}(z_t, t, f_{\text{atlas}}, f_j^{\text{real}})$ \Comment{Predict noise using VDM}\;
    $\mathcal{L} = \| \hat{\epsilon_t} - \epsilon_t \|_2^2$ \Comment{Compute MSE loss}\;
    $f_\text{atlas} \gets \texttt{Update}(f_\text{atlas}, \mathcal{L} )$ \Comment{Update garment atlas}\;
  }
\end{algorithm}

\section{Experiments}
    In this section, we first describe our datasets (\ref{ssec:datasets}) and evaluation metrics (\ref{ssec:metrics}). Then, we evaluate our method on image-to-360\degree\ (\ref{ssec:image-to-360}) and video-to-360\degree\ (\ref{ssec:video-to-360}) novel view synthesis (NVS), demonstrating quantitative and qualitative improvements over related methods (\ref{ssec:comparisons}) and ablated versions of our method (\ref{ssec:ablations}). We also include garment animation results in the supplementary material.

\subsection{Datasets}\label{ssec:datasets}
\vspace{-0.5em}
    We train our model using a combination of: (1) Real-world fashion images and videos:  17M crawled garment images and 52K garment videos. We additionally use the UBC Fashion Video dataset~\citep{ubc-dataset}, containing 500 train and 100 test videos. (2) Synthetic 3D garment assets: 8,473 unique garment assets, combination of turboquid~\citep{turbosquid}, objaverse~\citep{objaverse}, and online crawling. For each 3D garment, we render 32 views covering one full 360\degree\ orbit around the object center. Each image and video frame $I$ is preprocessed using an in-house equivalent of Graphonomy~\citep{graphonomy} to compute the corresponding 2D person keypoints $J_{2D}$, garment segmentation $I_g$, and 2D pose of the garment image $J_g$. Each 2D pose $J_{2D}$ is further preprocessed as a heat-map representation that is spatially-aligned with $I$.  Mediapipe~\citep{mediapipe} is used to compute 3D person keypoints $J_{3D}$. During evaluation, each input pair consists of 1 or 3 real-world images or frames form a held-out dataset similar to (1) and a randomly selected pose sequence from (2), covering one full 360\degree\ spin.
     
\begin{figure}[!tbp]
  \centering
  \includegraphics[width=\linewidth]{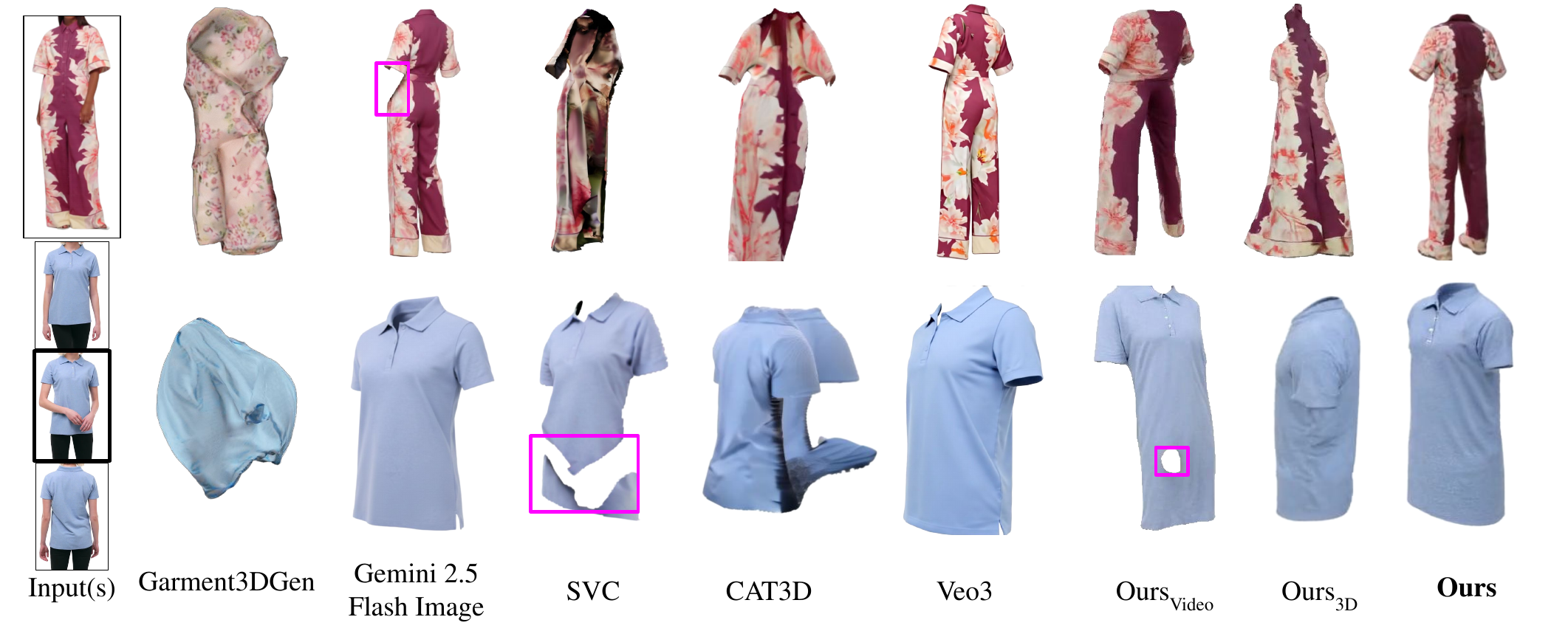}
  \caption{\textbf{Qualitative Comparisons.} 
  HoloGarment demonstrates superior preservation of garment appearance, realism, canonical pose, and multiview consistency, as well as robustness to occlusions, compared to related works and the ablated versions.}
  \label{fig:comparisons}
\end{figure}

\subsection{Evaluation Metrics}\label{ssec:metrics}
\vspace{-0.5em}
    We evaluate our method based on garment fidelity, multi-view consistency, and 3D realism. To measure garment fidelity, we compute the Fréchet Inception Distance FID~\citep{FID} and CLIP~\citep{CLIP} scores between the input garment images and the predicted garment images. To evaluate multi-view consistency and overall similarity to ground-truth 3D garments, we compute the Fréchet Video Distance (FVD)~\citep{FVD} and structural similarity (SSIM) between ground-truth rendered 360\degree\ spin videos and the output frames. 


\subsection{Garment Image-to-360\degree}\label{ssec:image-to-360}
\vspace{-0.5em}
    We showcase qualitative results from our image-to-360\degree\ NVS method in Figure~\ref{fig:qualitative}. HoloGarment synthesizes consistent and realistic novel views in a canonical target pose from single-view (top row) or multi-view inputs (middle row), even in the presence of occlusions, pose changes, and wrinkling in the input images. Moreover, HoloGarment handles a variety of garments, including tops, dresses, jackets, rompers, and pants. 
    

    

\setlength{\tabcolsep}{5pt}

\begin{table}
\begin{center}
  \label{tab:comparisons}
  \begin{tabular}{c|ccccccccc}
    \toprule
    & \multicolumn{4}{c}{Custom Dataset} & \multicolumn{4}{c}{UBC Dataset} \\
    \midrule
    Method & FID $\downarrow$ & CLIP $\uparrow$ & FVD $\downarrow$ & SSIM $\uparrow$ & FID $\downarrow$ & CLIP $\uparrow$ & FVD $\downarrow$ & SSIM $\uparrow$ \\
    \midrule
    Garment3DGen & 320 & 0.631 & 2477 & 0.002 & 310 & 0.638 & 2534 & 1.72e-5 \\
    Gemini 2.5 Flash Image$^1$ & 156 & 0.836 & -- & 0.700 & 137 & 0.880 & -- & 0.742 \\
    SVC & 130 & 0.855 & 1109 & 0.767 & 200 & 0.835 & 1144 & 0.745 \\
    CAT3D & 152 & 0.861 & 1073 & 0.578  & 186 & 0.798 & 1137 & 0.625\\
    $\text{ours}_{\text{video}}$ & 131 & \bf{0.890} & 1088 & 0.730 & 128 & 0.872 & 1053 & 0.743 \\
    $\text{ours}_{\text{3D}}$ & 144 & 0.871 & 968 & 0.739 & 147 & 0.847 & \bf{880} & 0.754  \\
    $\bf{ours}$ & \bf{128} & 0.872 & \bf{875} & \bf{0.729} & \bf{127} & \bf{0.881} & \bf{880} & \bf{0.771}\\
  \bottomrule
\end{tabular}
\caption{\textbf{Quantitative Comparisons.} HoloGarment outperforms Gemini 2.5 Flash Image~\citep{Google_Gemini_2_5_Flash_image}, Stable Virtual Camera (SVC)~\citep{stable_virtual_camera}, Garment3DGen~\citep{garment3dgen}, and CAT3D~\citep{cat3d} on all metrics. Our method also achieves competitive results on both garment fidelity (FID, CLIP) and 3d garment realism (FVD, SSIM) metrics when compared to video-only and 3D-only training. $^1$Given that the Gemini 2.5 Flash Image model isn't designed for temporal consistency, we chose to omit the FVD metric from our evaluation.}
\end{center}
\vspace{-2em}
\end{table}

\subsection{Garment Video-to-360\degree}\label{ssec:video-to-360}
\begin{wrapfigure}{r}{0.5\textwidth}
  \centering
  \vspace{-1.5em}
  \includegraphics[width=\linewidth]{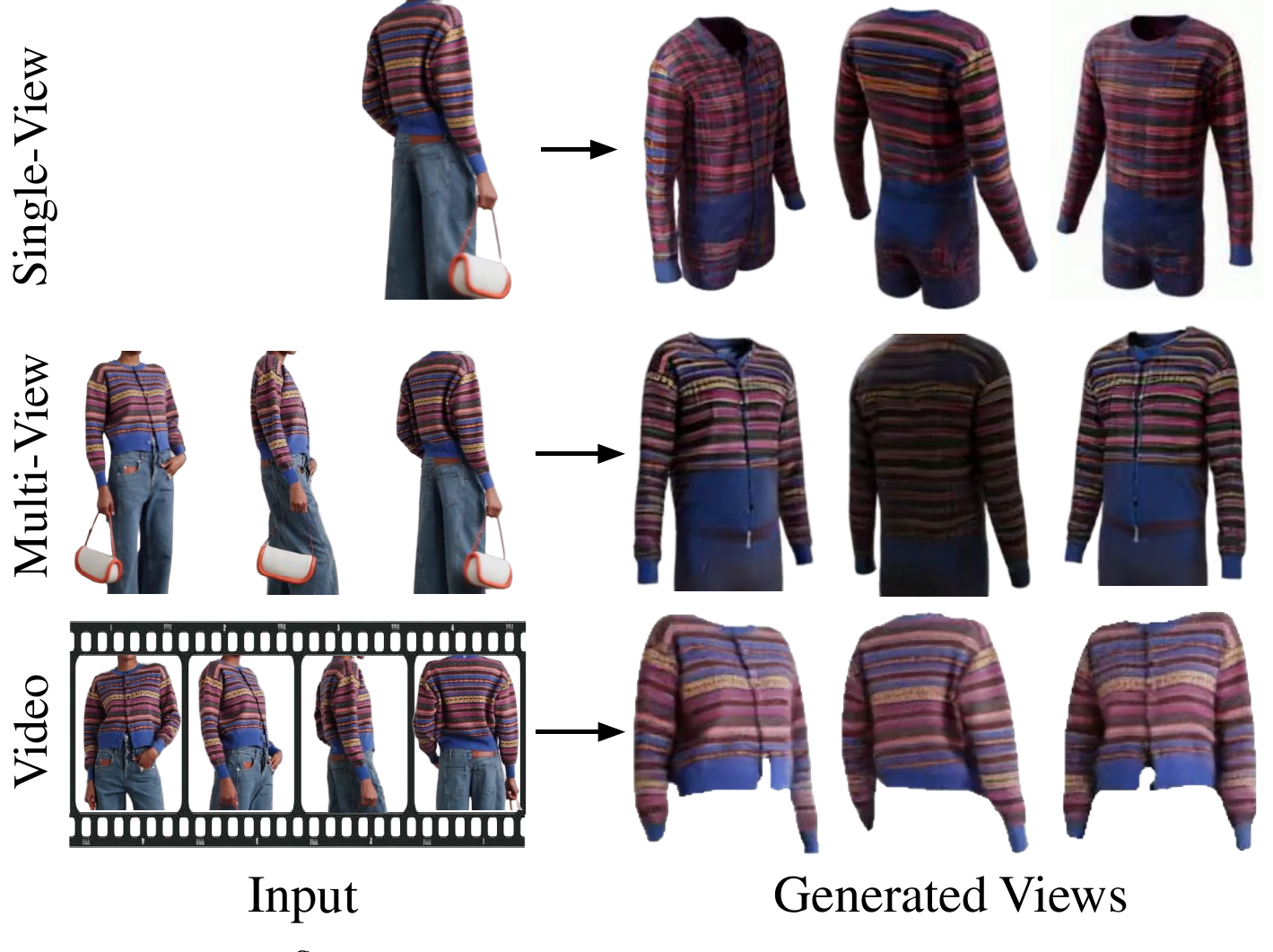}
  \caption{\textbf{Atlas Finetuning Ablation.} In single-view (\textit{top row}) and multi-view conditioning (\textit{middle row}), poorly chosen input views negatively affect the quality of synthesized views. Atlas finetuning on video (\textit{bottom row}) eliminates the dependency on input view selection by consolidating details from all video frames to improve garment texture details and multi-view consistency.}
  \label{fig:atlas-ablation}
  \vspace{-1.3em}
\end{wrapfigure}

\begin{figure}[ht]
  \centering
  \includegraphics[width=1\linewidth]{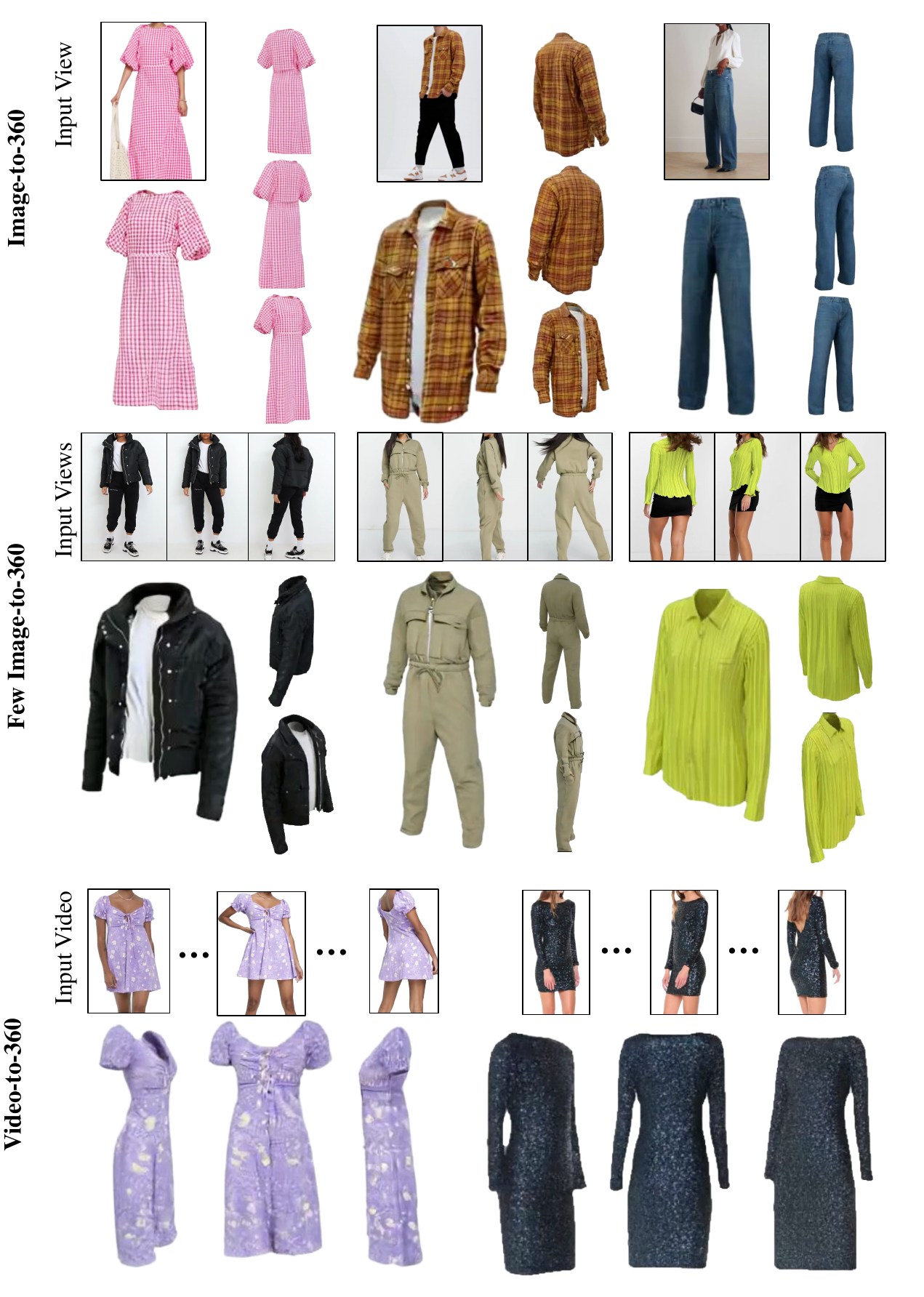}
  \caption{\textbf{Qualitative Results.} Our method generates 360\degree\ novel views of garments from single images, multiple images, or videos. Additional qualitative results are shown in the supplementary material. }
  \label{fig:qualitative}
\end{figure}

\vspace{-0.5em}
    To synthesize 3D-consistent 360\degree\ novel views of a real-world garment in a dynamic video, we finetune a latent garment embedding, which we call a garment ``atlas" $f_{\text{atlas}}$, on a specific 128-frame real-world video for 500 iterations with batch size 32 and constant learning rate of $1e{-3}$. By only optimizing for $f_{\text{atlas}}$ and activating only the video-specific temporal blocks of our model $\theta_{\text{spin}}$, we prevent undesired overfitting to the motion of the input video. 
    We showcase qualitative examples of our video-to-360\degree\ NVS method in Figure~\ref{fig:atlas-ablation} and the bottom row of Figure~\ref{fig:qualitative}. Our atlas finetuning strategy enables HoloGarment to consolidate an arbitrary number garment views, poses, and deformations into a unified 360\degree\ garment representation. 

    

\subsection{Comparisons to State-of-the-Art}\label{ssec:comparisons}
\vspace{-0.5em}
    We compare our method to image-to-3D (Garment3DGen~\citep{garment3dgen} and CAT3D~\citep{cat3d}), image editing (Gemini 2.5 Flash Image~\citep{Google_Gemini_2_5_Flash_image}), and camera-controlled video generation (Stable Virtual Camera~\citep{stable_virtual_camera}, Veo3~\citep{veo3}) methods. Our results are presented in Figure~\ref{fig:comparisons} and Table~1.
    
    Qualitatively, HoloGarment produces superior results compared to Garment3DGen, CAT3D, and Stable Virtual Camera, and achieves visual quality on par with large, publicly available models like Gemini 2.5 Flash Image and Veo3. This is notable given that our model was trained on a significantly smaller dataset.
    
    Quantitatively, our approach consistently outperforms all compared methods across all metrics on both evaluation datasets (Table 1). Due to the high computational cost of large-scale evaluation, a full quantitative comparison with Veo3 is reserved for future work. We also omit the FVD metric for Gemini 2.5 Flash Image since it is not designed for temporal consistency. Further discussion and implementation details for each method are provided in the supplementary material.

\subsection{Ablations}\label{ssec:ablations}
    \textbf{Implicit Training Paradigm:} In Figure~\ref{fig:comparisons} and Table 1, we demonstrate the benefit of our implicit video-and-3D training approach. We compare our pretrained base model trained only on video data ($\text{ours}_{\text{video}}$), trained only on synthetic 3D data ($\text{ours}_{\text{3D}}$), and jointly trained on video and 3D data ($\text{ours}$). While training our model on video data alone leads to the great photorealism and garment fidelity, it fails to enable 3D-consistent motion generation and realistic novel views.  Plus, the synthesized views contain holes where limbs or hair overlap with the garment, because real video data has occlusions that lead to such holes after segmentation. On the other hand, with 3D-only training, our model generates highly consistent and realistic garment spins. However, this model exhibits poor garment fidelity and tends to oversmooth textures and patterns, due to the limited size and diversity of the 3D dataset. Our jointly-trained model balances the benefits of video training and 3D training: it maintains garment fidelity and photorealism, while simultaneously generating consistent, canonical novel views without holes.
    
    \begin{wraptable}{r}{0.5\linewidth}
  \centering
  \label{tab:atlas-ablation}
  \setlength{\tabcolsep}{0.5em}
  \vspace{-1em}
  \begin{tabular}{cccccc}
    \toprule
    Method & FID $\downarrow$ & CLIP $\uparrow$ & FVD $\downarrow$ & SSIM $\uparrow$ \\
    \midrule
    $\text{w/o Atlas}$ & 134 & 0.852 & 538 & 0.689 \\
    $\text{w/ Atlas}$ & \textbf{103} & \textbf{0.900} & \textbf{474} &  \textbf{0.728}  \\
  \bottomrule
\end{tabular}
\caption{\textbf{Atlas Finetuning Ablations.} We quantitatively compare 40 results with and without atlas finetuning. Non-atlas results were conditioned on a single input view.}
\vspace{-1em}
\end{wraptable}

    \textbf{Atlas Finetuning:} We qualitatively and quantitatively evaluate our atlas finetuning strategy. Figure~\ref{fig:atlas-ablation} demonstrates that a poorly-chosen input view drastically limits novel view realism. While few-view conditioning improves fidelity, it is still limited by the quality of the input views. On the other hand, atlas finetuning enables our model to consolidate information from an arbitrary number of images, improving fidelity and realism. This finding is supported quantitatively in Table 2. Compared to single-image conditioning, atlas finetuning improves performance across all metrics.

\section{Discussion}

In this paper, we present HoloGarment, a method for synthesizing state-of-the-art novel views of garments in real-world images and videos. We introduce an implicit training scheme to optimize a video diffusion model for real-world garment image-to-360\degree\ novel-view synthesis (NVS) using a combination of large-scale 2D garment data and limited synthetic 3D garment assets. We further propose atlas finetuning, a strategy where a garment embedding, or ``atlas", is finetuned on a dynamic garment video to enable video-to-NVS capabilities. 


\textbf{Limitations:} While our method improves over existing methods, it faces several limitations. Due to the limited diversity of the synthetic 3D garment dataset, HoloGarment struggles with unusual garment shapes (e.g. assymmetry or cut-outs). Our model also exhibits some bias towards those garment categories which are more abundant in the 3D dataset, such as pants and t-shirts. See the supplementary for qualitative examples. A larger synthetic garment dataset may remedy such issues.  Other future work includes speeding up atlas finetuning (currently $\sim$30 minutes on a single TPU) and increasing resolution via super-resolution network. 


\clearpage
\clearpage

\bibliography{main}
\bibliographystyle{iclr2025_conference}

\clearpage

\appendix

\textbf{\Large Supplementary Material}

\section{Architecture Details}\label{ssec:supp-architecture}
We show our overall video diffusion model (VDM) architecture in Figure~\ref{fig:architecture}. Our model is trained following v-prediction~\citep{v_prediction}, where the model outputs the predicted noise $\epsilon'_t$ at timestep $t$, such that $z'_{t-1} = z_{t} + \epsilon'_t$. The $L2$-loss is computed in $\epsilon$ space.

\textbf{Input Preprocessing:} The conditional inputs to our VDM are 1 or 3 segmented garment images $I_g$, their corresponding 2D poses $J_g$, and a sequence of driving 2D and 3D poses $(J_{2D}, J_{3D})$. In the case of multiple input garment images, $I_g$ is channel-wise concatenation of the segmented garment images. Additionally, $I_g$ and its corresponding spatially-aligned 2D poses $J_g^{2D}$ are concatenated channel-wise. 

\textbf{Conditioning Inputs:} The noisy video $z_t$, garment signals $[I_g, J_g]$, and driving 2D poses $J_{2D}$ are encoded by separate UNet encoders~\citep{tryondiffusion} into features $f_z$, $f_g$, $f_{j2D}$, respectively. The driving 3D poses $J_{3D}$ are separately encoded by 4 dense layers into features $f_{j3D}$ and reshaped to match $f_{j2D}$. The conditioning input embeddings $f_g, f_{j2d}, f_{j3d}$ are processed by the DiT blocks~\citep{DiT} before the UNet decoder. 
We concatenate the driving pose features, $f_{j2D}$ and $f_{j3d}$, with the noisy video features $f_z$. The input garment image and pose features $f_g$ are cross-attended with noisy video features $f_z$, in order to implicitly warp the input garment features to their target locations according to the driving poses~\citep{tryondiffusion}, getting warped features $f_z'$.

\textbf{UNet:} Similarly to Fashion-VDM~\citep{fashion-vdm}, we add 3D convolution, temporal attention, and temporal mixing blocks after the two lowest-resolution spatial layers in the UNet encoder and decoder. However, unlike Fashion-VDM, we duplicate these temporal blocks, such that one set only processes video batches and the other only processes 3D spin batches. We implement this switch in the network via conditional network branching. Note that only one branch of temporal blocks is activated at a time. Finally, a UNet decoder decodes $f_z'$ into the predicted noise $\epsilon_{t}$. 

Additional implementation details of our architecture include:
\begin{itemize}
    \item The kernel sizees of our Conv2D and Conv3D blocks are $(3, 3)$ and $(4, 3, 3)$, respectively.
    \item Our 8 DiT blocks are implemented with 8 attention heads each and a hidden size of 2048 at feature resolution $32$x$24$.
    \item Our UNet encoders consist of 4 downsampling CNN blocks. Symmetrically, our UNet decoders consist of 4 upsampling CNN blocks.
\end{itemize}

\begin{figure}[!ht]
  \centering
  \includegraphics[width=\linewidth]{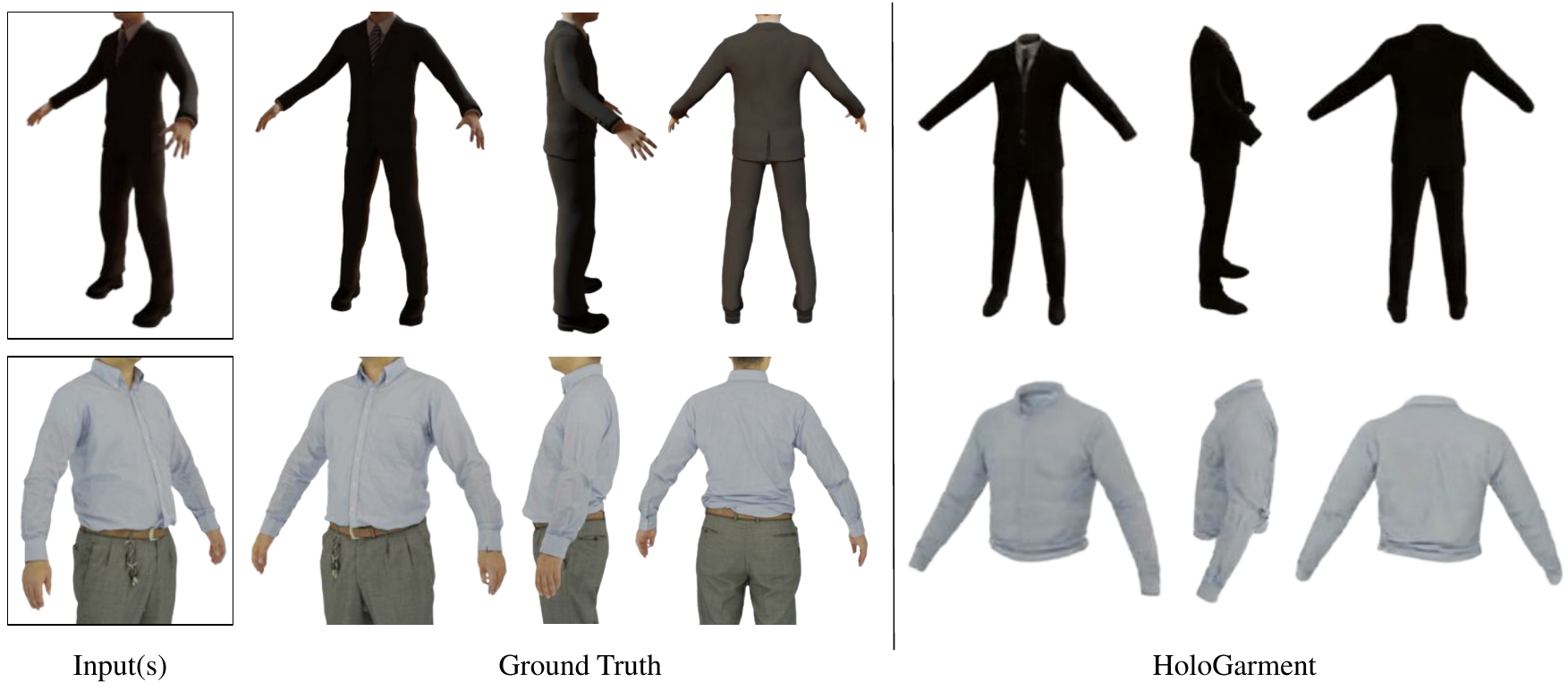}
  \caption{\textbf{Qualitative Results on Synthetic 3D Garments.} HoloGarment generates consistent 360-degree novel views from a single garment view that retain high-fidelity to their corresponding ground-truth synthetic 3D assets.}
  \label{fig:gt_recon_3d}
\end{figure}

\section{Training and Inference Details}\label{ssec:implementation-details} 
    
    \textbf{Training:} We pretrain our base model without temporal layers on our custom garment image dataset for 1M iterations with batch size 8~\citep{fashion-vdm}. Then, we train our full model with temporal layers for 589K iterations, approximately 3 days, on 16 TPU-v4's. In this stage of training, we train jointly with 33\% fashion image data, 33\% fashion video data, and 33\% novel views rendered from 3D garment assets, each with frame count $32$ and resolution 512 x 384. For task (2) batches, 3D garment assets are rendered as 360-degree RGB spin videos, similar to \citep{cat3d}. However, different from \citep{cat3d}, which uses camera position as conditioning, we compute $J_{2D}$ and $J_{3D}$ for both garment spins and real fashion videos, so that the motion representation is shared. Plus, 2D and 3D together encapsulate camera pose~\citep{PnP}. Finally, we finetune the model for an additional 50K iterations on 3D data only.
    
    For both pretraining and training, we use an Adam optimizer with linearly decaying learning rate of $1e{-4}$ to $1e{-5}$ over a maximum of 1M iterations. We additionally add independent dropout for each conditional input with 10\% probability per batch. After each forward pass, we compute the L2 loss on predicted noise $\epsilon_t$ at diffusion timestep $t$.

    \textbf{Inference:} During inference, we use the DDPM sampler \citep{ddpm} with 1000 refinement steps to generate 32-frame videos. For evaluating our full and ablated models, we employ dual classifier-free guidance~\citep{instruct-pix2pix} with conditioning groups ($\emptyset$, $I_g$, $[J_{2D}, J_{3D}]$) and weights $(1, 5, 1)$.

\begin{figure}[!tbp]
  \centering
  \includegraphics[width=0.8\linewidth]{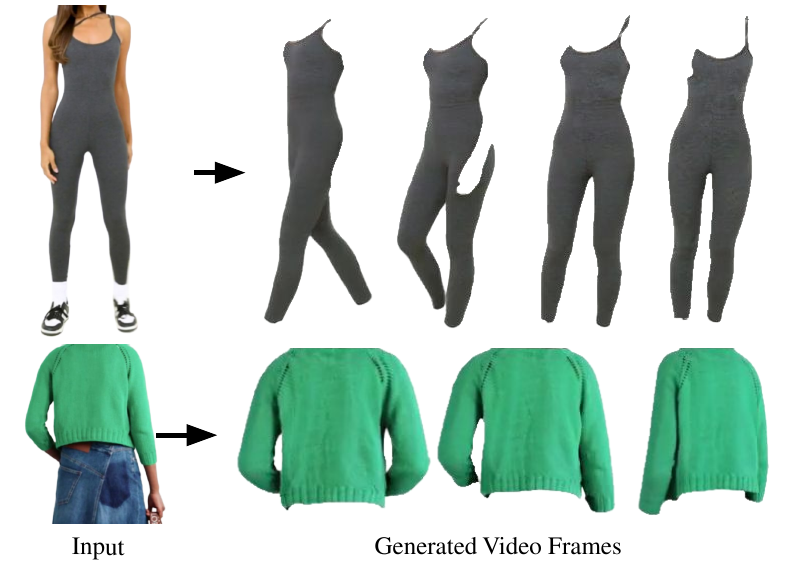}
  \caption{\textbf{Qualitative Animation Results.} HoloGarment generates realistic garment animations given a garment image and dynamic driving pose sequence. }
  \label{fig:animation}
  \vspace{-1.5em}
\end{figure}

\section{Comparisons to State-of-the-Art Details}

    \textbf{Gemini 2.5 Flash Image:} We compare our method to Gemini 2.5 Flash Image~\citep{Google_Gemini_2_5_Flash_image} on garment novel view synthesis. For quantitative evaluation, we generate the front view of each input segmented garment using the prompt \textit{"Generate a front-facing image of the garment in a-pose and without occlusions."} As Gemini 2.5 Flash Image is not designed for temporal consistency, we omit FVD in the quantitative comparison. For qualitative comparisons, we specify the target output angle of the garment in the prompt. Despite synthesizing high-quality, plausible novel views, Gemini 2.5 Flash Image is not designed for temporally-consistency, and struggles to generate smooth 360\degree\ novel views.

    \textbf{Veo3:} We qualitatively compare with Veo3~\citep{veo3} frame-to-video functionality through the Flow app of Google labs using the input segmented garment image as the input frame and the prompt \textit{``Generate a 360-degree orbit of this garment in a-pose and without occlusions."} Veo3 generates high-quality, consistent orbits, but over-saturates the garment colors.
    
    \textbf{Stable Virtual Camera:} We evaluate the official implementation of Stable Virtual Camera (SVC)~\citep{stable_virtual_camera} provided by the authors. We run SVC in single-image video generation mode on the input segmented garment image, following an orbit trajectory for 32 target frames and using all other default parameters. In Figure~\ref{fig:comparisons}, SVC fails to inpaint occluded regions (bottom row), synthesize plausible novel views (top row) or generate the garment in a canonical a-pose.
    
    \textbf{Garment3DGen:} We follow the official Garment3DGen implementation. Garment3DGen does not provide a texturing tool, so we use a text-to-texture model~\citep{flashtex}, as suggested by the authors. We caption each input image using Gemini~\citep{gemini}, then use FlashTex~\citep{flashtex} to add texture to the 3D mesh. From the caption, we also determine the garment type and select the closest template mesh provided by the authors. As shown in Figure~\ref{fig:comparisons}, the requirement of a template mesh severely limits Garment3DGen's ability to generalize to different garment shapes. Plus, the generated textures show poor fidelity to the input garment image. In contrast, HoloGarment is independent of any template meshes or third-party texture generation methods, making it robust to diverse garment shapes and textures.
    
    \textbf{CAT3D:} For comparisons with CAT3D, we follow the original authors' single-image implementation. Figure~\ref{fig:comparisons} shows that CAT3D generates a flattened appearance in side and back views. Additionally, as shown in the bottom row, CAT3D is not robust to occlusions or pose variations, reproducing the input holes and wrinkling. Finally, unlike our method, CAT3D cannot warp the garment into a desired target pose, replicating the original garment pose. 
    


\section{Additional Qualitative Results}\label{ssec:supp-qualitative-results}

\begin{figure}[h]
  \centering
  \includegraphics[width=\linewidth]{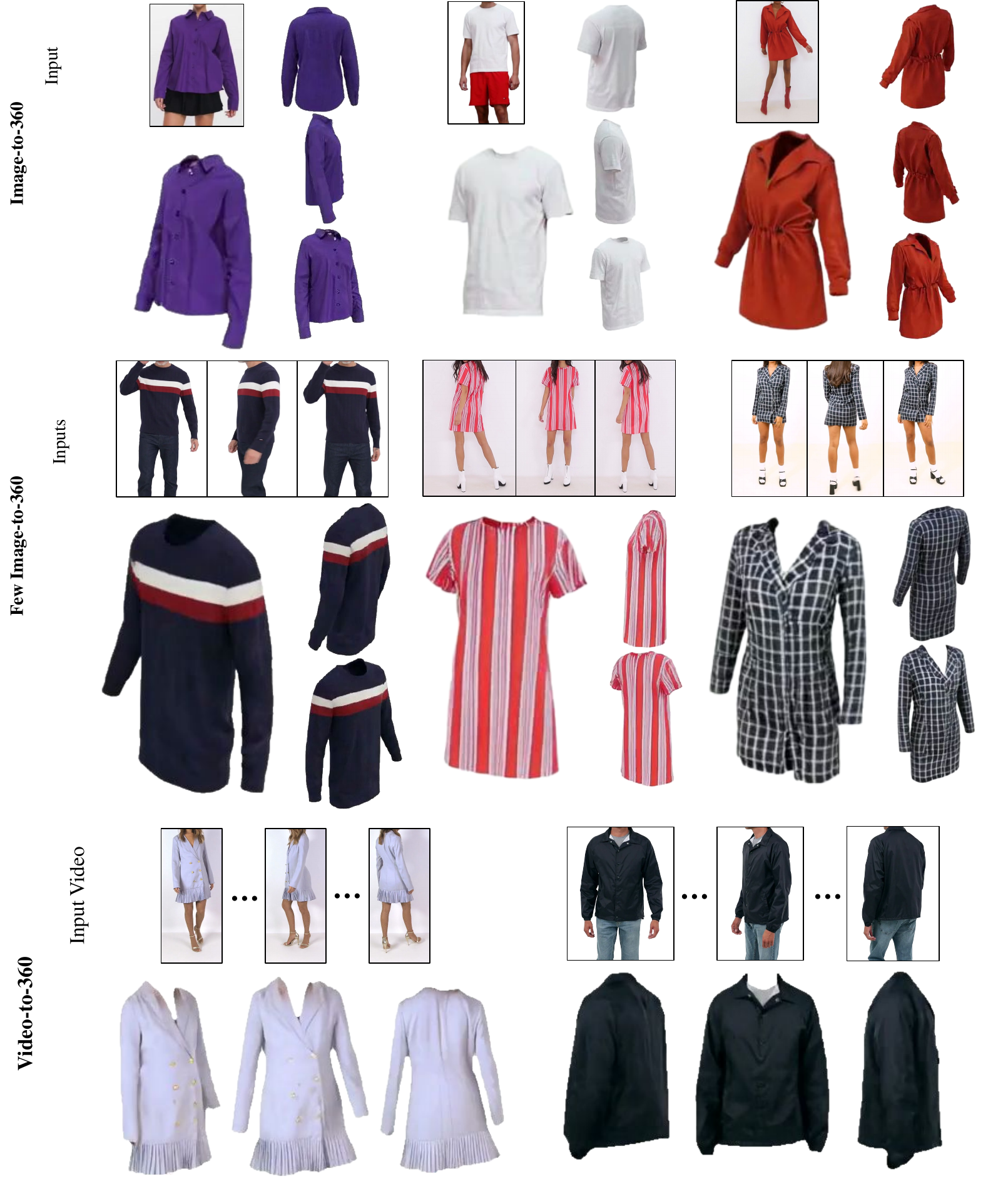}
  \caption{\textbf{Additional Qualitative Results.} Demonstrating our method's capability for 360\degree\ novel view synthesis of garments from a 1-3 input image(s) or a video. Results are best viewed in our supplementary video.}
  \label{fig:supp_qualitative}
\end{figure}

We show additional qualitative image-to-3D, sparse view-to-3D, and video-to-3D results of our method in Figure~\ref{fig:supp_qualitative}. We also show qualitative results of our single-view method on \begin{wrapfigure}{r}{0.5\textwidth}
  \includegraphics[width=\linewidth]{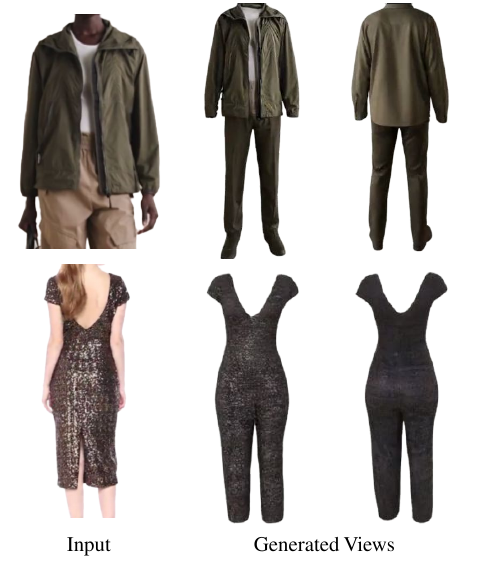}
  \caption{\textbf{Failure Cases.} At times, HoloGarment hallucinates pants when the input garment is top-only ($\textit{left}$). In other cases, due to ambiguities in the input (e.g., dress slits, extremely large occlusions), it may generate an incorrect garment shape ($\textit{right}$). }
\label{fig:failure-cases}
\end{wrapfigure} held-out synthetic 3D garment assets in Figure~\ref{fig:gt_recon_3d}. Given a single view of a synthetic 3D asset, HoloGarment synthesizes plausible novel views that are consistent and retain high fidelity to the ground truth views.


\section{Garment Animation}
In Figure~\ref{fig:animation}, we demonstrate HoloGarment's ability to realistically animate real-world garments given an image and driving pose sequence (task 1). Due to the nature of the real-world image and video data, HoloGarment creates wrinkling and occlusions when operating on video data. Although the focus of our work is NVS, the ability for HoloGarment to perform well on the image animation task, is crucial for enabling video-to-NVS finetuning (Section~\ref{ssec:video-to-360}).




\section{Failure Cases}
  
In Figure~\ref{fig:failure-cases}, we show two examples of failure cases of our method. These include hallucinating bottoms for top-only garment inputs and at times misrepresenting skirts as pants, especially when a slit is present in the input view.

\clearpage
\pagebreak

\end{document}